\pdfoutput=1

\documentclass[11pt]{article}

\usepackage[final]{acl}

\usepackage{times}
\usepackage{color}
\usepackage{amsmath} 
\usepackage{placeins}
\usepackage{geometry}
\usepackage{xcolor}
\usepackage{tcolorbox}
\tcbuselibrary{skins}
\usepackage{latexsym}
\usepackage{placeins}
\usepackage{listings}
\usepackage{enumitem}
\usepackage{url}
\usepackage{multirow}
\usepackage{rotating}
\usepackage{array}
\usepackage{booktabs}
\usepackage[T1]{fontenc}
\usepackage[utf8]{inputenc}
\usepackage[nopatch]{microtype}
\usepackage{setspace}

\usepackage{inconsolata}
\usepackage{graphicx}

\DeclareRobustCommand{\indicWords}[1]{%
  \raisebox{-\dp\strutbox}{%
    \includegraphics[page=\csname indicWords#1\endcsname]{indicWords}%
  }%
}

\newtcolorbox{promptbox}[1][]{
  enhanced,
  colback=lightgray,
  colframe=darkgray,
  fonttitle=\small\bfseries\sffamily\color{white},
  coltitle=white,
  attach boxed title to top left={yshift=-2mm, xshift=5mm},
  boxed title style={colback=darkgray},
  #1
}
\definecolor{orange}{rgb}{1,0.5,0}
\definecolor{darkgray}{RGB}{60,60,60}
\definecolor{lightgray}{RGB}{240,240,240}
\definecolor{sectioncolor}{RGB}{31, 78, 121}
\definecolor{answercolor}{RGB}{0, 102, 51}
\definecolor{justificationcolor}{RGB}{102, 102, 102}

\newtcolorbox{checklistitem}[2][]{
    colback=gray!5,
    colframe=gray!30,
    boxrule=0.5pt,
    arc=3pt,
    title={\textbf{#2}},
    fonttitle=\color{sectioncolor}\bfseries,
    top=8pt,
    bottom=8pt,
    left=8pt,
    right=8pt,
    toptitle=4pt,
    bottomtitle=2pt,
    before skip=8pt,
    after skip=8pt,
    #1
}

\newcommand{\Answer}[1]{\textbf{\color{answercolor}Answer:} #1}
\newcommand{\Justification}[1]{\textbf{\color{justificationcolor}Section or Justification:} #1}

\title{COMI-LINGUA: Expert Annotated Large-Scale Dataset for Multitask NLP in Hindi-English Code-Mixing}

\author{
 \textbf{Rajvee Sheth},
 \textbf{Himanshu Beniwal},
 \textbf{Mayank Singh}
\\
 LINGO Research Group, Indian Institute of Technology Gandhinagar, India
\\
 \small{
   \textbf{Correspondence:} \href{mailto:lingo@iitgn.ac.in}{lingo@iitgn.ac.in}
 }
}

\begin{document}
\maketitle
\begin{abstract}

We introduce COMI-LINGUA, the largest manually annotated Hindi-English code-mixed dataset, comprising 125K+ high-quality instances across five core NLP tasks: Token-level Language Identification, Matrix Language Identification, Named Entity Recognition, Part-Of-Speech Tagging and Machine Translation. Each instance is annotated by three bilingual annotators, yielding over 376K expert annotations with strong inter-annotator agreement (Fleiss' Kappa $\geq$ 0.81). The rigorously preprocessed and filtered dataset covers both Devanagari and Roman scripts and spans diverse domains, ensuring real-world linguistic coverage. Evaluation reveals that closed-weight LLMs significantly outperform traditional tools and open-weight models in zero-shot settings. Notably, one-shot prompting consistently boosts performance across tasks, especially in structure-sensitive predictions like POS and NER. Fine-tuning open-weight LLMs on COMI-LINGUA demonstrates substantial improvements, achieving up to 95.25 F1 in NER, 98.77 F1 in MLI, and competitive MT performance, setting new benchmarks for Hinglish code-mixed text. COMI-LINGUA is publicly available at this URL\footnote{\url{https://huggingface.co/datasets/LingoIITGN/COMI-LINGUA} Version 1.0, Updated till 15\textsuperscript{nd} September 2025.}.
\end{abstract}

\section{Introduction}
Code-mixing is the blending of multiple languages within a single utterance—a pervasive phenomenon in multilingual societies, especially on social media platforms \citep{jamatia2020deep, srivastava-singh-2020-phinc}. Over half of the world’s population is bilingual or multilingual and frequently uses mixed-language expressions in digital communication \citep{grosjean_2021}. In the Indian context, Hindi-English (Hinglish) code-mixed  text is particularly widespread and presents significant computational challenges due to orthographic complexity, frequent language switches, and script variation between Devanagari and Roman forms \citep{bali-etal-2014-borrowing,takawane2023language, 8554413}. A characteristic example is:
{\color{orange}Kal mujhe} \indicWords{Asentence}{\color{orange}hai}, {\color{blue}but} \indicWords{Bsentence}{\color{blue}will be an issue},
where Hindi and English tokens co-occur and certain English words like ``{\color{blue}office}'' and ``{\color{blue}traffic}'' may appear in Devanagari script.
(English Translation: “Tomorrow I have to go to the office, but traffic will be an issue.”)

Despite growing interest, current Hinglish datasets have critical limitations:
\textit{(1)} a predominant focus on Roman script, ignoring natural script variation \citep{begum-etal-2016-functions, bali-etal-2014-borrowing, srivastava2020understanding},
\textit{(2)} limited scale and coverage \citep{srivastava2021challenges, kumar-etal-2018-aggression, tiwari2024large, kartik2024synthetic},
\textit{(3)} in\-suf\-fi\-cient task diversity \citep{aguilar-etal-2020-lince, khanujaa2020new, bohra-etal-2018-dataset, khanuja-etal-2020-gluecos}, and
\textit{(4)} reliance on synthetic data generation and labeling rather than human annotation \citep{chatterjee2022pacman, srivastava-singh-2021-quality, kartik2024synthetic, sravani-mamidi-2023-enhancing}.

\begin{figure*}[!tbh]
\centering
\includegraphics[width=1\linewidth]{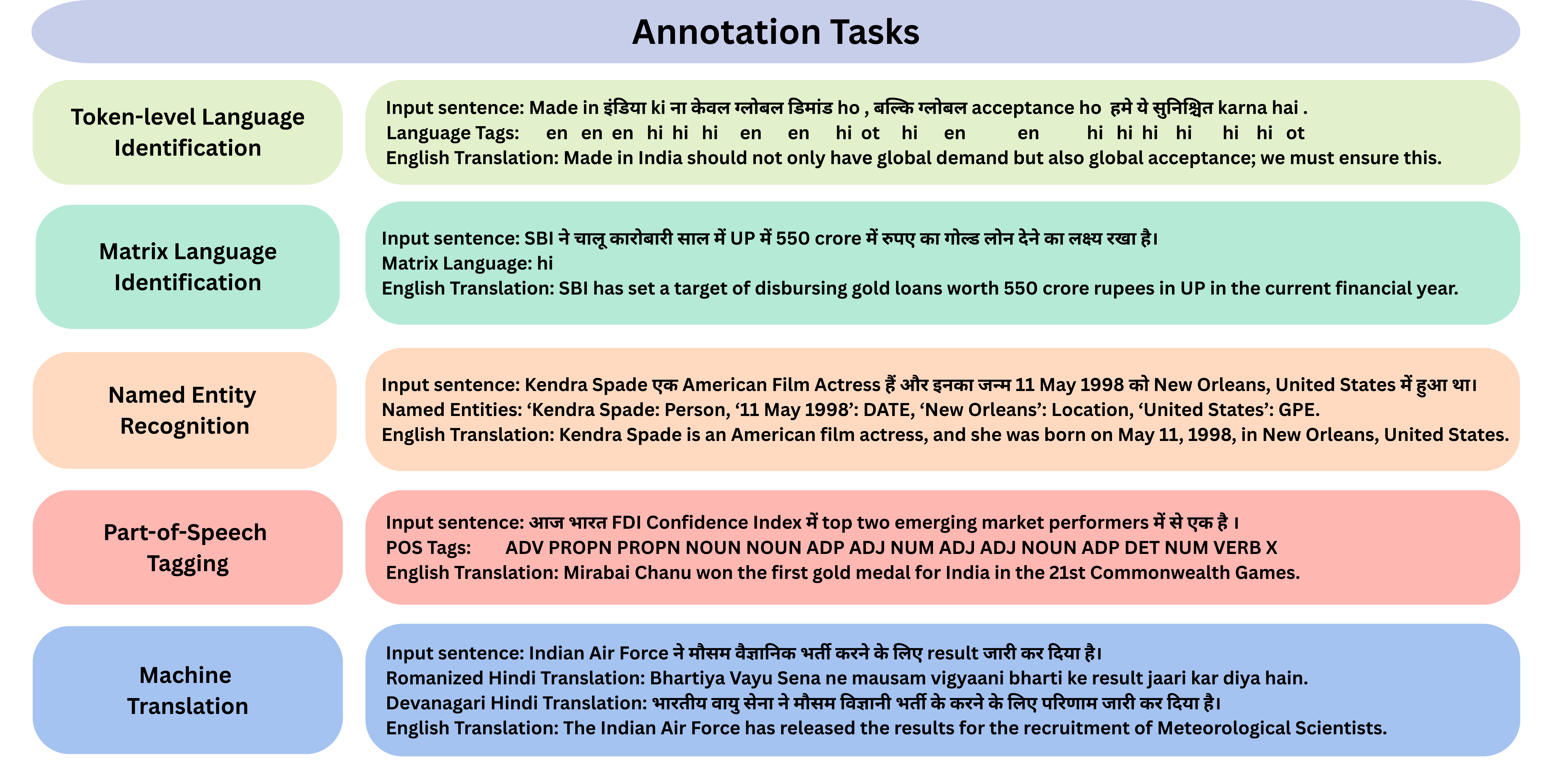}

\caption{Sample Annotations Across COMI-LINGUA Tasks: Shown here are annotated instances for each of the five tasks defined in the COMI-LINGUA task set, emphasizing the annotation strategy and linguistic diversity.}
\label{fig:annotation-tasks}
\end{figure*}

\noindent To address these limitations, we present a novel comprehensive dataset \textbf{COMI-LINGUA} (\textbf{CO}de-\textbf{MI}xing and \textbf{LING}uistic Insights on Natural Hinglish \textbf{U}sage and \textbf{A}nnotation) that advances Hinglish code-mixing research. The key contributions include:
\begin{itemize}
    \setlength{\itemsep}{0pt}
    \item Curation of the largest publicly available Hinglish dataset (376K manually annotated instances), released under a CC-BY-4.0 license, capturing real-world code-mixing behavior across both Roman and Devanagari scripts. Each instance is annotated by one annotator across one of the key NLP tasks: token-level language identification, matrix language identification, named entity recognition, part-of-speech tagging and machine translation.
    \item Robust benchmarking of state-of-the-art multilingual LLMs (mLLMs), including both open-weight and closed-weight models—alongside traditional NLP tools, under two inference paradigms: zero-shot and one-shot in-context learning.
    \item In-depth error analysis of mLLMs on code-mixed tasks, uncovering critical limitations such as misclassification of English borrowings in Devanagari script, context truncation, overfitting in one-shot settings, prompt mimicry, repetitive or hallucinated outputs, and practical deployment barriers like API usage constraints—highlighting persistent challenges in script-aware and context-sensitive language modeling. 
\end{itemize}

\section{Related Work}
\label{sec:rel_work}
Code-mixing—the blending of multiple languages in a single utterance—poses major challenges for NLP due to its structural variability \citep{srivastava2021challenges}. This is especially true for Hinglish, given their distinct scripts and syntax \citep{bali-etal-2014-borrowing}. Progress is hindered by the lack of large, annotated datasets, as collecting and labeling such data remains costly and labor-intensive \citep{srivastava2021challenges}.

\noindent \textbf{Language Identification}
is a foundational task in code-mixed NLP. Multiple approaches have been developed to detect language boundaries within mixed-language sequences, including statistical models, CRFs, and deep learning-based techniques \citep{Shekhar2020LanguageIF, singh-etal-2018-language, gundapu-mamidi-2018-word, molina-etal-2016-overview}. These efforts have paved the way for improved preprocessing and downstream modeling of code-mixed data.

\noindent \textbf{Named Entity Recognition} in code-mixed text has seen significant progress through both resource development and model improvements. \citet{dowlagar2022cmnerone} showed that leveraging multilingual data enhances NER accuracy, while \citet{ansari2019cross} created cross-script datasets using Wikipedia. Transformer-based approaches and meta-embeddings have also been effective in improving NER for Indian code-mixed data \citep{9074379}.

\noindent \textbf{Part-of-Speech Tagging}
A variety of annotated datasets have been introduced for POS tagging in code-mixed contexts. \citet{singh-etal-2018-twitter} and \citet{vyas-etal-2014-pos} developed corpora from Twitter and Facebook, respectively, while \citet{pratapa-etal-2018-word} generated synthetic datasets for evaluating bilingual word embeddings. \citet{sequiera-etal-2015-pos} experimented with various machine learning algorithms, and \citet{chatterjee2022pacman} introduced PACMAN, a large-scale synthetic POS-tagged dataset that achieved state-of-the-art performance in code-mixed POS tagging tasks.

\noindent \textbf{Machine Translation} for code-mixed content remains a growing research area. \citet{dhar-etal-2018-enabling} and \citet{srivastava-singh-2020-phinc} developed parallel corpora for Hinglish code-mixed sentences, while \citet{hegde-lakshmaiah-2022-mucs} proposed translation models using transliteration and pseudo-translation, achieving competitive results in the MixMT shared task at WMT 2022.

\noindent \textbf{Benchmarking and Evaluation Frameworks}
Several benchmark datasets have been introduced to evaluate NLP systems on code-mixed tasks. LinCE \citep{aguilar-etal-2020-lince} provides a comprehensive benchmark covering 11 corpora and 4 language pairs. GLUECoS \citep{khanuja-etal-2020-gluecos} demonstrated the benefits of fine-tuning multilingual models on code-switched datasets across multiple tasks. Emotion and sentiment annotation efforts, such as the Hinglish Twitter corpus by \citet{vijay-etal-2018-corpus}, the L3Cube-HingCorpus \citep{nayak2022l3cube}, and the emotion-annotated SentiMix dataset by \citet{ghosh2023multitasking} further support affective computing in code-mixed settings.

\noindent Despite ongoing efforts, standardized benchmarks for evaluating LLMs on diverse Hinglish code-mixed tasks—such as acceptability judgments, syntactic fluency, and translation fidelity—remain limited. Existing benchmarks are often narrow in scope and rely on synthetic or small-scale data. To address this, we curate the largest high-quality, human-annotated dataset for training and evaluating LLMs on a broad range of Hinglish code-mixed phenomena. It serves as both an evaluation suite
and a diagnostic tool to advance multilingual and code-mixed language understanding research.

\section{The COMI-LINGUA dataset}

\subsection{Raw Dataset Curation} \label{subsec:raw-dataset}
\label{sec:raw_dataset_curation}
We curated raw data from publicly accessible and licensed platforms spanning diverse categories such as news, politics, entertainment, social events, sports, and informational content, with a focus on the Indian subcontinent. Sources included prominent news portals and official digital archives, detailed in Appendix \S\ref{sec:appendix-sources}. The collected content was cleaned using regex-based preprocessing to remove noise such as advertisements, HTML tags, and footers, and then segmented into individual sentences. A Code-Mixing Index (CMI, \citet{das2014identifying}) was computed for each sentence, and only those sentences with a CMI score $\geq$ 9 were retained to ensure a substantial degree of code-mixing. Given the under-representation of mixed Devanagari-Roman script samples in existing datasets, we also collected supplementary data to enhance coverage and linguistic diversity. This includes enriching the dataset by incorporating additional Hinglish code-mixed samples from prior works~\citep{srivastava-singh-2020-phinc, gupta-etal-2023-mutant, singh-etal-2018-named} and from Hugging-Face\footnote{\url{https://huggingface.co/datasets/pardeep/youtube-vidoes-transcripts-hindi-english/}}. 

\subsection{Dataset Processing} \label{subsec:dataset-pre}
The preprocessing pipeline was designed to enhance the quality and neutrality of the corpus through rigorous noise reduction techniques. To ensure the dataset was both clean and relevant, we removed duplicate instances, hate speech, and abusive content. Sentences containing offensive or inappropriate language were identified and filtered out using established profanity and hate speech detection tools, including \textit{thisandagain}\footnote{\url{https://github.com/thisandagain/washyourmouthoutwithsoap/blob/develop/README.md}} and \textit{Hate-Speech-Detection-in-Hindi}\footnote{\url{https://github.com/victorknox/Hate-Speech-Detection-in-Hindi/blob/main/README.md}}.

At the token level, additional preprocessing steps were applied. Sentences with fewer than five tokens were discarded to eliminate non-informative content such as fragments, abbreviations, emojis, and filler phrases—commonly arising from typing errors or social media discourse. Examples of such removed content include: \textit{`\#GuessTheSong'}, \textit{`during dinner'}, and \textit{`@enlightenedme bas ek hi'}. Further data refinement was conducted during the manual annotation process (see Section~\ref{sec:data_refinement} for more details). 

\subsection{Data Annotation} \label{subsec:data-ann}
\label{sec:data_tasks}
To annotate the Hinglish code-mixed corpus, we employed \textsc{Commentator} \citep{sheth-etal-2024-commentator}, a robust annotation framework specifically designed for multilingual code-mixed text.

The annotation was carried out by a team of three graduate-level experts proficient in both Hindi and English. All annotators possess prior experience with social media content and demonstrate strong programming capabilities, along with familiarity with using version control systems. These competencies contributed to a systematic, efficient, and reproducible annotation process. The annotators were recruited specifically for this project and were compensated at a rate of approximately \$1.64 per hour. The funding for the annotation work was provided through a government-sponsored initiative; the compensation adheres to standard remuneration practices considered appropriate for the annotators’ qualifications and the economic context of their country of residence.

We selected five diverse annotation tasks, balancing well-established tasks with high reliability and underexplored challenges. Annotators followed detailed guidelines with examples to ensure consistency and clarity across tasks (Appendix \S\ref{sec:guidelines}, Figure~\ref{fig:annotation-tasks}). The tasks are: 

\begin{enumerate}
    \item \textbf{\textit{Token-level Language Identification (LID):}}
In this task, each token in the dataset was assigned one of three possible language labels: English (\textcolor{blue}{en}), Hindi (\textcolor{orange}{hi}), or Other (ot). Initial language tags were generated using Microsoft's Language Identification Tool\footnote{\url{https://github.com/microsoft/LID-tool}}, which served as a baseline for further manual refinement. As shown in Figure~\ref{fig:annotation-tasks}, each token is assigned a language tag.
    
  \item \textbf{\textit{Matrix Language Identification (MLI):}}
Each sentence is annotated with a Matrix Language, which identifies the dominant language governing the grammatical structure of the sentence. In code-mixed text, even when multiple languages are interspersed, one language typically dictates the syntactic and morphosyntactic framework of the utterance. Figure~\ref{fig:annotation-tasks} showcases a sentence annotated with its matrix language.
    
    \item \textbf{\textit{Named Entity Recognition (NER):}} In the NER task, each token in a sentence is annotated with a label from a predefined set of entity types outlined in Table~\ref{tab:entities}. These include conventional categories, such as Person, Location, Organization, Date/Time, and GPE (Geo-Political Entities), as well as social media-specific types like Hashtags, Mentions, and emojis. An instance of annotated entities across different types is shown in Figure~\ref{fig:annotation-tasks}. This allows the annotation schema to comprehensively capture the diversity and informality observed in code-mixed social media text.
   
  \begin{table}[t]
    \centering
    \small
    \begin{tabular}{ll}
    \toprule
    \textbf{Entity Type} & \textbf{Description} \\ 
    \midrule
    \textbf{Person} & Names of individuals \\ 
    \textbf{Location} & Non-political physical locations \\ 
    \textbf{Organization} & Institutions or companies \\ 
    \textbf{Date/Time} & Temporal expressions (e.g., dates) \\ 
    \textbf{GPE} & Geo-Political Entities \\ 
    \textbf{Hashtags} & Words prefixed by `\#' \\ 
    \textbf{Mentions} & User mentions prefixed by `@' \\ 
    \textbf{Emoji} & Emoticons conveying emotions \\ 
    \bottomrule
    \end{tabular}
    \caption{Named entity types and their descriptions in our annotation schema.}
    \label{tab:entities}
\end{table}

    \item \textbf{\textit{Part-of-Speech (POS) Tagging:}}
Each token in the code-mixed dataset was annotated with a Part-of-Speech (POS) tag selected from the \textbf{Universal POS tagset} proposed by \citet{singh-etal-2018-twitter}. The tagset, summarized in Table~\ref{tab:pos_tags}, was chosen for its language-agnostic design, enabling consistent annotation of Hindi and English words in a single sentence—an essential feature for handling code-mixed content effectively. A representative example is presented in Figure \ref{fig:annotation-tasks}. Initial predictions for POS tags were generated using the \textit{CodeSwitch NLP} library\footnote{\url{https://github.com/sagorbrur/codeswitch}}, which supports multilingual code-mixed data and provides pre-trained models suitable for tagging noisy, informal text commonly found on social media platforms. 

\begin{table}[t]
    \centering
    \small
    \begin{tabular}{@{\hspace{0.3cm}}p{1.3cm}@{\hspace{0.8cm}}p{4.5cm}}
    \toprule
    \textbf{POS Tag} & \textbf{Description} \\ 
    \midrule
    \textbf{NOUN} & Common nouns \\ 
    \textbf{PROPN} & Proper nouns \\ 
    \textbf{VERB} & Verbs in all tenses and moods \\ 
    \textbf{ADJ} & Adjectives describing nouns \\ 
    \textbf{ADV} & Adverbs modifying verbs \\ 
    \textbf{ADP} & Adpositions \\ 
    \textbf{PRON} & Pronouns \\ 
    \textbf{DET} & Determiners \\ 
    \textbf{CONJ} & Conjunctions \\ 
    \textbf{PART} & Grammatical particles \\ 
    \textbf{PRON\_WH} & Wh-pronouns \\ 
    \textbf{PART\_NEG} & Negative particles \\ 
    \textbf{NUM} & Numerals and cardinal numbers \\ 
    \textbf{X} & Typos, abbreviations, punctuation \\ 
    \bottomrule
    \end{tabular}
    \caption{Part-of-speech tags and their descriptions used in our annotation schema.}
    \label{tab:pos_tags}
\end{table}
    
    \item \textbf{\textit{Machine Translation (MT):}}
This task involves constructing parallel translations for code-mixed sentences into three distinct formats: (i)  Standard English, (ii) Romanized Hindi and (iii) Devanagari Hindi. The goal is to facilitate a multilingual Hinglish sentence to align with its respective translations across scripts and languages.
A representative translation instance across the three formats is shown in Figure~\ref{fig:annotation-tasks}. Initial translation predictions were generated using the LLaMA 3.3 language model\footnote{\url{https://github.com/meta-llama/llama-models/blob/main/models/llama3_3/MODEL_CARD.md}}.
    
\end{enumerate}

\noindent For all tasks, we used state-of-the-art NLP tools or LLMs for automated pre-annotation, generating initial labels based on task-specific criteria. Expert annotators then refined these outputs through manual post-editing. This two-stage process ensured high-quality annotations while improving consistency and speeding up dataset creation.

\subsection{Manual Data Refinement} \label{subsec:dataset-refine}
\label{sec:data_refinement}
During the annotation phase, the dataset underwent iterative refinement to ensure quality and consistency, guided by annotator feedback on instances to be excluded (see Table~\ref{tab:annotation-examples} in Appendix \S\ref{sec:appendix-examples}). We removed sentences if they (i) were monolingual English or Hindi, (ii) lacked relevant linguistic tags or named entities, contained no meaningful content, or merged multiple instances into one, or (iii) included languages other than Hindi and English, which were beyond the scope of this study. This refinement process was crucial for preserving corpus integrity and ensuring that the final dataset consisted solely of high-quality Hinglish code-mixed text. The Raw and Filtered columns in Table~\ref{tab:annotation_analysis} represent the number of original instances provided for initial annotation and the final number of instances retained after annotation, respectively. The difference between these values corresponds to instances flagged by annotators as not satisfying the manual annotation criteria.

\subsection{Annotation Efforts and Quality} \label{subsec:annotation}
\label{sec:annotation_quality}

The manual annotation process involved substantial human effort across all tasks, particularly in refining the outputs of automated tools. For example, for the LID task, each annotator reviewed 504,102 tokens and flipped an average of 95,670 tokens—approximately 19\% of the original predictions. In the POS task, 63,002 of 427,941 tokens were corrected, indicating a 15\% flip rate. Similarly, for the NER task, each annotator modified about 98,760 out of 538,160 tokens, translating to 18\% manual corrections.  For the MLI task, no initial predictions were provided, leading to 100\% of the sentences being annotated. 
To assess annotation reliability, we computed inter-annotator agreement (IAA) using Fleiss' Kappa \citep{fleiss1971measuring}, a standard metric for evaluating consistency among multiple annotators on categorical labels \citep{Hallgrenandkevin}. All classification tasks achieved Fleiss' Kappa scores above \textbf{0.817}, indicating substantial to near-perfect agreement (Table~\ref{tab:annotation_analysis}). As machine translation is a generative task, IAA was not calculated. While not a direct measure of quality, the final dataset retains a high level of code-mixing, with an average CMI exceeding \textbf{14} across tasks, ensuring strong code-mixing.

\begin{table}[t]
\centering
\small

\renewcommand{\arraystretch}{1.0}
\begin{tabular}{lcccc}
\toprule
\textbf{Task} & \textbf{Raw} & \textbf{Filtered} & \textbf{IAA} & \textbf{CMI} \\
\midrule
\textbf{LID}  & 29,950 & 25,772 & 0.834 & 20.87 \\
\textbf{MLI}  & 29,950 & 25,772 & 0.976 & 20.87 \\
\textbf{NER}  & 26,929 & 24,913 & 0.852 & 14.38 \\
\textbf{POS}  & 27,229 & 24,598 & 0.817 & 21.60 \\
\textbf{MT} & 26,727 & 24,558 & - & 17.07 \\
\midrule
\textbf{Total / Avg.} & 140,785 & 125,615 & 0.863 & 18.96 \\
\bottomrule
\end{tabular}
\caption{Corpus Statistics: The Raw and Filtered columns represent the number of original instances provided for initial annotation and the final instances retained after annotation, respectively. Note: IAA was not computed for the MT task as it is a generative task.}
\label{tab:annotation_analysis}
\end{table}

\begin{table*}[t]
\centering
\small
\setlength{\tabcolsep}{0.75em} 
\renewcommand{\arraystretch}{1.2}
\begin{tabular}{lcccccccccccc}
\toprule
\textbf{Model/Library} & \multicolumn{3}{c}{\textbf{LID}} & \multicolumn{3}{c}{\textbf{MLI}} & \multicolumn{3}{c}{\textbf{NER}} & \multicolumn{3}{c}{\textbf{POS}} \\
 & $P$ & $R$ & $F_1$ & $P$ & $R$ & $F_1$ & $P$ & $R$ & $F_1$ & $P$ & $R$ & $F_1$ \\
\midrule
\multicolumn{13}{c}{\textbf{Zero-shot}} \\
\midrule
\texttt{claude-3.5-sonnet} & 92.8 & 92.4 & 92.1 & 98.8 & 83.5 & 90.0 & 59.1 & 55.1 & 56.7 & 75.3 & 64.8 & 69.0 \\
\texttt{gpt-4o} & \textbf{92.8} & \textbf{92.8} & \textbf{92.7} & 98.4 & 97.9 & 98.1 & 60.5 & 60.1 & 60.1 & \textbf{76.1} & \textbf{66.0} & \textbf{70.1} \\
\texttt{gemini-1.5-Flash} & 82.9 & 40.4 & 47.9 & 98.8 & 21.4 & 33.7 & 44.2 & 44.2 & 43.8 & 73.4 & 62.4 & 66.5 \\
\texttt{LLaMA-3.3-instruct} & 73.4 & 73.7 & 73.3 & 98.8 & 59.0 & 73.1 & \textbf{67.5} & \textbf{67.3} & \textbf{66.8} & 74.3 & 65.5 & 68.9 \\
\texttt{mistral-instruct} & 54.5 & 39.0 & 42.4 & 98.1 & 58.7 & 72.3 & 65.1 & 41.5 & 50.2 & 10.2 & 6.72 & 7.78 \\
\texttt{command-a-03-2025} & 92.0 & 92.0 & 91.8 & \textbf{98.5} & \textbf{98.0} & \textbf{98.3} & 65.9 & 67.8 & 66.6 & 73.5 & 65.4 & 68.6 \\
\texttt{codeswitch} & - & - & - & - & - & - & 81.6 & 83.1 & 81.2 & \textbf{89.1} & \textbf{87.8} & \textbf{88.2} \\
\texttt{Microsoft LID} & 80.2 & 76.5 & 74.4 & - & - & - & - & - & - & - & - & - \\
\midrule
\multicolumn{13}{c}{\textbf{One-shot}} \\
\midrule
\texttt{claude-3.5-sonnet} & 93.0 & 92.7 & 92.5 & \textbf{98.8} & \textbf{98.9} & \textbf{98.8} & \textbf{85.9} & \textbf{85.2} & \textbf{85.0} & 81.4 & 79.2 & 79.3 \\
\texttt{gpt-4o} & \textbf{93.9} & \textbf{94.0} & \textbf{93.8} & 98.7 & 97.7 & 98.1 & 77.4 & 75.8 & 76.0 & 81.6 & 78.0 & 78.9 \\
\texttt{gemini-1.5-Flash} & 80.2 & 76.5 & 74.4 & 98.4 & 40.4 & 56.4 & 66.5 & 67.5 & 66.0 & 72.9 & 64.6 & 68.0 \\
\texttt{LLaMA-3.3-instruct} & 90.3 & 89.6 & 89.3 & 98.8 & 97.8 & 98.2 & 79.0 & 79.1 & 78.4 & \textbf{85.1} & \textbf{84.0} & \textbf{84.1} \\
\texttt{mistral-instruct} & 72.1 & 70.0 & 70.1 & 98.3 & 88.1 & 92.7 & 65.5 & 44.4 & 52.6 & 77.3 & 66.9 & 69.8 \\
\texttt{command-a-03-2025} & 92.1 & 91.7 & 91.3 & 98.9 & 98.7 & 98.3 & 76.7 & 78.9 & 77.3 & 74.5 & 65.7 & 69.5 \\
\bottomrule
\end{tabular}
\caption{Performance metrics on the COMI-LINGUA test sets for various models across different experimental settings (Zero-shot, One-shot) and tasks: LID, MLI, NER, and POS tagging. $P$, $R$, and $F_1$ denote Precision, Recall, and F1-score respectively. `-' indicates that the task is not supported by the respective tool.}
\label{tab:performance_metrics_all}
\end{table*}

\begin{table}[t]
\centering
\small
\setlength{\tabcolsep}{0.3em} 
\renewcommand{\arraystretch}{0.9}
\begin{tabular}{lcccccc}
\toprule
\textbf{Model} & \multicolumn{2}{c}{\textbf{En}} & \multicolumn{2}{c}{\textbf{RH}} & \multicolumn{2}{c}{\textbf{DH}} \\
 & $B_{en}$ & $cF_{en}$ & $B_{rh}$ & $cF_{rh}$ & $B_{dh}$ & $cF_{dh}$ \\
\midrule
\multicolumn{7}{c}{\textbf{Zero-shot}} \\
\midrule
\texttt{claude-3.5-sonnet} & 48.1 & 63.5 & 48.6 & 64.4 & 56.0 & 65.7 \\
\texttt{gpt-4o} & 28.8 & 42.4 & 27.5 & 41.7 & 32.0 & 41.8 \\
\texttt{gemini-1.5-Flash} & 48.2 & 66.0 & 28.9 & 50.4 & 56.9 & 66.8 \\
\texttt{LLaMA-3.3-instruct} & \textbf{55.4} & \textbf{71.4} & \textbf{50.5} & \textbf{68.1} & \textbf{59.8} & \textbf{71.5} \\
\texttt{mistral-instruct} & 23.5 & 49.9 & 5.4 & 25.3 & 18.1 & 40.7 \\
\texttt{command-a-03-2025} & 38.6 & 58.5 & 35.2 & 57.0 & 48.8 & 61.8 \\
\midrule
\multicolumn{7}{c}{\textbf{One-shot}} \\
\midrule
\texttt{claude-3.5-sonnet} & 50.9 & 68.5 & 52.2 & 69.4 & 63.5 & 73.4 \\
\texttt{gpt-4o} & 50.2 & 68.6 & 50.2 & 68.4 & 58.4 & 70.1 \\
\texttt{gemini-1.5-Flash} & 39.5 & 57.6 & 40.5 & 58.7 & 58.9 & 69.6 \\
\texttt{LLaMA-3.3-instruct} & \textbf{62.2} & \textbf{74.8} & \textbf{54.8} & \textbf{71.3} & \textbf{60.7} & \textbf{70.3} \\
\texttt{mistral-instruct} & 30.0 & 53.8 & 19.6 & 44.1 & 18.5 & 40.5 \\
\texttt{command-a-03-2025} & 52.9 & 68.6 & 42.0 & 60.4 & 56.1 & 66.6 \\
\bottomrule
\end{tabular}
\caption{MT performance metrics on the COMI-LINGUA test sets for various models across Zero-shot and One-shot settings. $B_{en}$, $B_{dh}$, and $B_{rh}$ represent BLEU scores and $cF_{en}$, $cF_{dh}$, and $cF_{rh}$ represent chrF++ scores for Standard English, Devanagari Hindi, and Romanized Hindi translation outputs respectively.}
\vspace{-\baselineskip}
\label{tab:mt_performance_metrics}
\end{table}

\begin{table*}[t]
\centering
\small
\setlength{\tabcolsep}{0.6em} 
\renewcommand{\arraystretch}{1.2}
\begin{tabular}{lcccccccccccc}
\toprule
\textbf{Model} & \multicolumn{3}{c}{\textbf{LID}} & \multicolumn{3}{c}{\textbf{MLI}} & \multicolumn{3}{c}{\textbf{NER}} & \multicolumn{3}{c}{\textbf{POS}} \\
 & $P$ & $R$ & $F_1$ & $P$ & $R$ & $F_1$ & $P$ & $R$ & $F_1$ & $P$ & $R$ & $F_1$ \\
\midrule
\multicolumn{13}{c}{\textbf{Zero-shot}} \\
\midrule
\texttt{LLaMA-3.1-8B-In} & \textbf{62.50} & \textbf{70.04} & \textbf{62.37} & 96.83 & 4.68 & 8.93 & \textbf{63.40} & \textbf{66.80} & \textbf{64.86} & 57.86 & 19.75 & 26.84 \\
\texttt{aya-expanse-8b} & 51.08 & 70.55 & 59.05 & \textbf{98.71} & \textbf{59.56} & \textbf{74.25} & 54.47 & 68.27 & 59.88 & \textbf{76.92} & \textbf{29.50} & \textbf{40.55} \\
\texttt{Qwen2.5-7B-In} & 56.43 & 69.63 & 59.13 & 98.64 & 21.20 & 34.82 & 61.02 & 65.36 & 57.37 & 68.40 & 8.70 & 9.06 \\
\midrule
\multicolumn{13}{c}{\textbf{One-shot}} \\
\midrule
\texttt{LLaMA-3.1-8B-In} & \textbf{83.03} & \textbf{80.30} & \textbf{81.44} & 98.55 & 59.52 & 74.16 & 72.51 & 67.35 & 68.54 & \textbf{72.72} & \textbf{63.22} & \textbf{64.73} \\
\texttt{aya-expanse-8b} & 73.03 & 71.07 & 70.48 & \textbf{98.35} & \textbf{81.36} & \textbf{89.00} & \textbf{79.73} & \textbf{81.44} & \textbf{79.18} & 55.29 & 48.70 & 48.20 \\
\texttt{Qwen2.5-7B-In} & 57.17 & 74.85 & 64.83 & 98.49 & 61.94 & 75.74 & 74.18 & 76.30 & 74.04 & 70.46 & 59.89 & 63.09 \\
\midrule
\multicolumn{13}{c}{\textbf{Fine-tuned}} \\
\midrule
\texttt{LLaMA-3.1-8B-In} & \textbf{95.29} & \textbf{94.57} & \textbf{94.75} & 98.07 & 88.76 & 93.05 & \textbf{95.28} & \textbf{95.24} & \textbf{95.25} & 86.66 & 86.16 & 86.17 \\
\texttt{aya-expanse-8b} & 87.45 & 86.92 & 87.15 & \textbf{98.69} & \textbf{98.90} & \textbf{98.77} & 94.94 & 94.91 & 94.90 & \textbf{88.97} & \textbf{88.55} & \textbf{88.61} \\
\texttt{Qwen2.5-7B-In} & 84.44 & 80.23 & 81.74 & 97.82 & 93.86 & 95.80 & 94.33 & 94.31 & 94.27 & 89.01 & 88.53 & 88.60 \\
\bottomrule
\end{tabular}
\caption{Performance metrics on the COMI-LINGUA test sets for three LLMs across different experimental settings (Zero-shot, One-shot, Fine-tuned) on four tasks: LID, MLI, NER, and POS tagging. Metrics shown are Precision ($P$), Recall ($R$), and F1-score ($F_1$). Abbreviations \texttt{LLaMA-3.1-8B-In} and \texttt{Qwen2.5-7B-In} denote \texttt{LLaMA-3.1-8B-Instruct} and \texttt{Qwen2.5-7B-Instruct} respectively.}
\label{tab:performance_fine_matrix}
\end{table*}

\noindent The \textbf{COMI-LINGUA} consists of 125,615 high-quality instances spanning five tasks, each independently annotated by three expert annotators, yielding a total of 376,845 annotations
(see Table \ref{tab:dataset_comparison}). To our knowledge, it is the largest manually annotated code-mixed dataset to date. For each task,
We provide two random splits: a test set of 5,000 instances and a training set comprising the remainder (as detailed in Table~\ref{tab:dataset_comparison}, Appendix \S\ref{sec:experiments}). Zero- and one-shot prompting was evaluated only on the fixed test set, whereas fine-tuning was carried out on the training split, with performance reported on the same 5,000-instance test set.  

\begin{table}[t]
\centering
\small
\setlength{\tabcolsep}{0.3em} 
\renewcommand{\arraystretch}{1.0}
\begin{tabular}{lcccccc}
\toprule
\textbf{Model} & \multicolumn{2}{c}{\textbf{En}} & \multicolumn{2}{c}{\textbf{RH}} & \multicolumn{2}{c}{\textbf{DH}} \\
 & $B_{en}$ & $cF_{en}$ & $B_{rh}$ & $cF_{rh}$ & $B_{dh}$ & $cF_{dh}$ \\
\midrule
\multicolumn{7}{c}{\textbf{Zero-shot}} \\
\midrule
\texttt{LLaMA-3.1-8B-In} & \textbf{38.3} & \textbf{67.5} & \textbf{15.6} & \textbf{49.2} & 7.4 & 13.5 \\
\texttt{aya-expanse-8b} & 33.2 & 67.1 & 4.80 & 20.0 & \textbf{25.6} & \textbf{57.5} \\
\texttt{Qwen2.5-7B-In} & 29.8 & 61.5 & 14.2 & 46.8 & 3.31 & 21.8 \\
\midrule
\multicolumn{7}{c}{\textbf{One-shot}} \\
\midrule
\texttt{LLaMA-3.1-8B-In} & \textbf{45.8} & \textbf{72.4} & \textbf{35.3} & \textbf{67.0} & 17.9 & 53.2 \\
\texttt{aya-expanse-8b} & 31.7 & 65.1 & 29.7 & 63.7 & 26.4 & 59.1 \\
\texttt{Qwen2.5-7B-In} & 30.2 & 61.7 & 18.3 & 52.8 & \textbf{35.6} & \textbf{60.7} \\
\midrule
\multicolumn{7}{c}{\textbf{Fine-tuned}} \\
\midrule
\texttt{LLaMA-3.1-8BIn} & \textbf{56.1} & \textbf{78.7} & \textbf{66.6} & \textbf{85.9} & \textbf{73.5} & \textbf{86.2} \\
\texttt{aya-expanse-8b} & 55.0 & 78.1 & 62.4 & 83.7 & 69.3 & 86.0 \\
\texttt{Qwen2.5-7B-In} & 51.9 & 76.0 & 63.9 & 84.1 & 63.8 & 78.4 \\
\bottomrule
\end{tabular}
\caption{MT performance on the COMI-LINGUA test set. $B_{en}$, $B_{rh}$, and $B_{dh}$ denote BLEU scores, while $cF_{en}$, $cF_{rh}$, and $cF_{dh}$ correspond to chrF++ scores for Standard English, Romanized Hindi, and Devanagari Hindi respectively. Abbreviations \texttt{LLaMA-3.1-8B-In} and \texttt{Qwen2.5-7B-In} denote \texttt{LLaMA-3.1-8B-Instruct} and \texttt{Qwen2.5-7B-Instruct} respectively.}
\vspace{-\baselineskip}
\label{tab:performance_mt}
\end{table}

\section{Experiments} \label{sec:exp}

\subsection{Baseline Tools and LLMs}
\label{sec:baselines}
We conducted a comprehensive evaluation of existing tools and language models on the COMI-LINGUA Benchmark. Our experimental setup spans traditional NLP toolkits, state-of-the-art open-weight LLMs, and proprietary commercial models. These systems are evaluated on their performance across five diverse Hinglish code-mixed NLP tasks, detailed in Section~\ref{sec:data_tasks}.

\noindent The traditional tools evaluated in this study include the \texttt{Microsoft LID}\footnote{\url{https://github.com/microsoft/LID-tool}} for token-level language identification and the \texttt{codeswitch} toolkit\footnote{\url{https://github.com/sagorbrur/codeswitch}} for POS and NER tasks in multilingual text, which provides a rule-based pipeline for annotating syntactic and semantic information in code-switched corpora. The four commercial closed-weight systems considered in our evaluation include : \texttt{claude-3.5-Sonnet} \citep{anthropic2024claude35sonnet}, \texttt{gpt-4o} \citep{achiam2023gpt}, \texttt{gemini-1.5-Flash} \citep{team2023gemini} and \texttt{command-a-03-2025} (111B) \citep{cohere2025command}. In addition, we assess open-weight LLMs \texttt{llama-3.3-instruct} (70B) \citep{touvron2023llama} and \texttt{mistral-instruct} (7B) \citep{jiang2023mistral}. 

We create specific prompt templates for each task to generate accurate, task-aligned responses from LLMs. The prompt template includes a high-level description of the task, specific annotation or tagging rules, and illustrative examples wherever applicable. For each of our five tasks, we developed two prompt variants: a zero-shot version providing only task instructions and a one-shot version that includes a single demonstrative example with instructions. The prompts are presented as a system-level instruction, followed by the user-supplied test input (i.e., a code-mixed sentence or token sequence). The complete prompt template used for each task under each prompt variant is detailed in Appendix \S\ref{sec:appendix-prompts}.

\subsection{Evaluation Metrics}
\label{sec:metrics}
We employ a suite of standard evaluation metrics, appropriately chosen for each task's nature. For token-level classification tasks—LID, POS, and NER—we report Precision (P), Recall (R), and the F\textsubscript{1}-score, computed at the macro level. For the MLI task, which is a sentence-level classification problem, we adopt the same classification metrics—P, R, and F\textsubscript{1}—computed on a per-sentence basis. For MT, we use the BLEU score~\citep{papineni2002bleu} and chrF++ score~\citep{popovic2015chrf} to evaluate the quality of translated outputs. Given the multilingual nature of our dataset, BLEU and chrF++ is computed separately for each output format: $B_{\text{en}}$, $cF_{\text{en}}$ for English, $B_{\text{rh}}$, $cF_{\text{rh}}$ for Romanized Hindi, and $B_{\text{dh}}$, $cF_{\text{dh}}$ for Devanagari Hindi. This disaggregated evaluation helps assess script-specific translation quality and is especially relevant given the transliteration variability in informal code-mixed text.

\subsection{Evaluation Configurations}
\label{sec:eval_configuration}
We evaluate model performance under three distinct paradigms: \textit{zero-shot} and \textit{one-shot} in-context learning, and task-specific \textit{fine-tuning}. Traditional NLP tools and libraries are inherently limited to zero-shot settings, as they rely on fixed rule-based or statistical models without the capability for contextual adaptation. In contrast, LLMs are evaluated under both zero-shot, one-shot and fine-tuned configurations to investigate their ability to generalize from instructions alone and to leverage minimal contextual supervision, respectively.

In the zero-shot setting, the prompt includes only task-specific instructions and formatting constraints without any illustrative examples. For the 1-shot setting, we augment the prompt with a single representative example demonstrating the input-output structure of the task. This example is carefully selected to reflect typical task behavior and is kept fixed across all evaluations to maintain consistency. For fine-tuning, we train models on task-specific training splits using formatted instruction-response pairs, allowing models to learn code-mixing patterns and task structures through parameter updates. Detailed illustrations of both prompt configurations for each task are provided in Appendix \S\ref{sec:appendix-prompts} and for fine-tuning, detailed hyperparameters are provided in Appendix \S\ref{sec:hyperparameters}.

\section{Results and Observations}
\label{sec:results}
Table~\ref{tab:performance_metrics_all} present the empirical results obtained under the two experimental configurations: zero-shot and one-shot in-context learning, respectively. It is important to note that traditional tools such as \texttt{codeswitch} and \texttt{Microsoft LID} are limited in their task coverage; results for tasks not supported by these tools are omitted from the tables.

\noindent \textbf{Traditional Tools vs. LLMs}:  The comparative analysis of traditional NLP tools and LLMs reveals clear distinctions in performance across code-mixed tasks. As shown in Table~\ref{tab:performance_metrics_all}, traditional tools such as \texttt{codeswitch} and \texttt{Microsoft LID} demonstrate strong performance on specific tasks they were designed for, particularly POS and LID, respectively. For instance, \texttt{codeswitch} achieves the highest POS F1 of 88.2, outperforming all LLMs in this task, while \texttt{Microsoft LID} attains a reasonable F1 of 74.4 for LID. However, these tools exhibit significant limitations in task coverage; they do not support MLI, MT, or tasks involving complex reasoning or generation.

\noindent \textbf{Open vs. Closed LLMs}
The performance gap between proprietary (closed) and open-weight LLMs is evident across both zero-shot and few-shot settings. In zero-shot mode, closed models such as \texttt{gpt-4o} and \texttt{claude-3.5-sonnet} dominate with top-tier results in most tasks. For example, gpt-4o achieves 92.7 F1 on LID and 98.1 F1 on MLI, while \texttt{claude-3.5-sonnet} reaches 92.1 F1 on LID and 90.0 F1 on MLI. However, when moving to a one-shot setting, open-weight models like \texttt{LLaMA-3.3-instruct} start closing the gap. Its performance improves significantly: LID F1 rises from 73.3 to 89.3, POS tagging reaches 84.1 (even surpassing \texttt{gpt-4o}), and NER climbs to 78.4. MT performance also peaks at 62.2 $B_{en}$ and 74.8 $cF_{en}$ for English, the highest across all models.

\noindent \textbf{Zero vs. One-shot Inference}
The transition from zero-shot to one-shot inference leads to notable performance improvements across most models and tasks. This is especially evident in complex tasks such as NER and MT, where providing one task-specific instance helps models disambiguate entities and manage code-mixed structures more effectively. For example, \texttt{claude-3.5-sonnet}'s NER F1 increases significantly from 56.7 in the zero-shot setting to 85.0 in the one-shot setting, while \texttt{LLaMA-3.3-instruct}'s $B_{en}$ improves from 55.4 to 62.2, alongside $cF_{en}$ scores increasing from 71.4 to 74.8 and $cF_{rh}$ from 68.1 to 71.3. \texttt{gpt-4o} similarly benefits, with NER performance rising from 60.5 to 77.4 and $B_{dh}$ improving from 32.0 to 58.4 and $cF_{dh}$ from 41.8 to 70.1. Open models like \texttt{LLaMA-3.3-instruct} also see considerable gains, such as POS tagging jumping from 68.9 to 84.1 and $cF_{en}$ MT reaching 74.8. These results demonstrate that even minimal supervision through a single example can significantly enhance model performance on linguistically complex, low-resource, or code-mixed tasks. At the same time, tasks like MLI exhibit relatively modest gains, suggesting that more deterministic tasks benefit less from one-shot prompting. Overall, one-shot inference provides a practical and effective method to unlock the latent capabilities of LLMs in multilingual and code-mixed scenarios. 

\noindent \textbf{Fine-Tuning LLMs with COMI-LINGUA} To further explore model performance beyond zero- and one-shot prompting, we fine-tuned the \texttt{LLaMA-3.1-8B-Instruct}, \texttt{aya-expanse-8b} and \texttt{Qwen2.5-7B-Instruct} models separately on each of the five COMI-LINGUA tasks using the respective training splits. Fine-tuning was carried out using task-specific formatted instructions, allowing the model to internalize both code-mixing patterns and structural nuances. Table~\ref{tab:performance_fine_matrix}
present the empirical results obtained under the three experimental configurations: zero-shot and one-shot in-context learning and fine-tuning respectively. The results demonstrate notable improvements across all tasks, with fine-tuned models outperforming traditional tools, open-weight baselines, and in some cases closed-weight LLMs.  

\noindent \textbf{Fine-tuning vs. Prompting Approaches} The fine-tuning results from Tables~\ref{tab:performance_fine_matrix} and~\ref{tab:performance_mt} demonstrate substantial performance gains over both zero-shot and one-shot inferencing approaches. Fine-tuned models consistently outperform traditional tools and achieve competitive or superior results compared to closed-weight LLMs across all tasks. Particularly notable improvements are observed in NER (30-40\% gains over prompting approaches) and consistent high performance in MLI (>95\% F1 across all fine-tuned models). In the MT task, the model yielded BLEU scores of 56.1 for $B_{en}$, 66.6 for $B_{rh}$ and 73.5 for $B_{dh}$. Correspondingly, the chrF++ scores were 78.7 for $cF_{en}$, 85.9 for $cF_{rh}$ and 86.2 for $cF_{dh}$ MT. 

These results highlight the strength of supervised fine-tuning on high-quality, diverse code-mixed data, as provided by COMI-LINGUA. Unlike zero- or one-shot setups, which depend heavily on prompt engineering and model prior knowledge, fine-tuning allows the model to generalize deeper linguistic patterns and task-specific strategies.

\section{Challenges with Current LLMs}\label{sec:challenges}
A consistent challenge across all models is the inability to accurately handle English borrowings written in Devanagari script—words like ``\indicWords{ksentence}'' and ``\indicWords{lsentence}'' were frequently misclassified as Hindi, reflecting a gap in script-aware language identification. Another prominent issue is sentence truncation; longer code-mixed inputs often lead to incomplete or abruptly cut-off outputs, indicating that many models struggle to preserve context over extended sequences. Models such as \texttt{gemini-1.5-flash} and \texttt{mistral-instruct} displayed repetitive generation patterns, producing redundant phrases within the same response. These models also occasionally injected subjective explanations into their outputs, despite clear instructions to extract objective information—for instance, adding interpretive statements when identifying the matrix language. Several models tended to mirror patterns from the prompt rather than perform actual analysis, indicating shallow understanding. Sentences with high grammatical or script variability posed yet another barrier, where many models, especially \texttt{gemini-1.5-flash} and \texttt{mistral-instruct}, failed to generate any output at all. Overfitting to examples also emerged as a concern, particularly in one-shot settings; models like \texttt{gpt-4o} and \texttt{command-a-03-2025} occasionally produced outputs that mimicked example structures rather than responding appropriately to the test input. This over-reliance was particularly evident in tasks such as MLI and LID, where one-shot performance slightly declined. Additionally, models hallucinated non-existent entities, suggesting overgeneralization from minimal supervision. (See Table~\ref{tab:llm-failure-cases} in the Appendix).

Beyond these general limitations, our analysis of smaller models (7--8B parameters) uncovered some failure patterns; for MT task, \texttt{Qwen2.5-7B-Instruct} inappropriately provided empty outputs with structured labels such as ``Unit 1: English, Unit 2: Romanized Hindi, Unit 3: Devanagari Hindi'' instead of providing actual translations. In NER tasks, \texttt{aya-expanse-8b} misclassified punctuation marks, tagging ``('' as Opening parenthesis and ``)'' as Closing parenthesis rather than using standard entity categories and labelling as `X'. POS tagging revealed more hallucination patterns, with \texttt{LLaMA-3.1-8B-Instruct} generating repetitive sequences like ``VERB NOUN PROPN NOUN NOUN NOUN NOUN'' for multiple instances. More concerning was the tendency of these models to output code snippets instead of task responses —\texttt{LLaMA-3.1-8B-Instruct} and \texttt{Qwen2.5-7B-Instruct} output Python import statements and function templates, rather than returning direct predictions as:
\begin{quote}
\texttt{import nltk\\
from nltk import pos\_tag\\
\# Download the required NLTK data \\
nltk.download(`perceptron\_tagger')\\
nltk.download(`punkt')}
\end{quote}
Similar code generation patterns emerged across tasks, with the model providing \texttt{import re} and \texttt{langdetect} modules for LID task rather than providing the actual labels. Entity hallucination was prevalent in \texttt{Qwen2.5-7B-Instruct}, which generated anomalous labels like (Live India) as `X X X X X X' and inappropriately tagged terms such as ``\indicWords{assentence}'' as HASHTAG entity. MT task suffered from incomplete generation, with outputs abruptly ending mid-sentence, as observed in Romanized Hindi Translation: ``Madras Hāikōrt ne dāk vibhāg ko.''(See Table~\ref{tab:8b-model-failures} in the Appendix). These systematic failures across all three smaller models highlight the importance of robust fine-tuning and careful prompt engineering when deploying compact LLMs for complex multilingual tasks.

\section{Conclusion and Future Directions}\label{sec:conclude}

LLMs often struggle with tasks like POS tagging, NER, and MT in code-mixed Hindi-English due to their lack of exposure to Indian multilingual data. Errors such as mislabeling entities or hallucinating content arise from limited training on structurally complex and script-variable inputs. The COMI-LINGUA dataset addresses these issues by providing high-quality, task-diverse, and richly annotated code-mixed text. Fine-tuning on this dataset enables models to better handle linguistic ambiguity, reduce overfitting, and improve reliability across tasks. Its inclusion of contextual examples and diverse sources—like social media and news—enhances the models’ ability to generalize across formal and informal registers, while iterative refinement through active learning ensures sustained performance gains.

\label{sec:bibtex}

\section*{Limitations}

While this study offers valuable insights into the annotation and processing of Hinglish code-mixed text, several limitations warrant consideration:
\begin{enumerate}
    \setlength{\itemsep}{0pt}
    \item \textbf{Language Pair Specificity:} The findings derived from Hinglish code-mixed data may not generalize to other language pairs (e.g., Spanish-English), given differences in syntactic structure, sociolinguistic norms, and code-switching behavior.
    
    \item \textbf{Demographic Bias:} The use of a relatively small and homogeneous group of annotators may introduce demographic bias, potentially limiting the broader applicability and reliability of the acceptability ratings.
    
    \item \textbf{Resource Constraints:} Scaling this work to other code-mixed language pairs remains challenging due to the scarcity of high-quality annotated corpora and the limited availability of models capable of robustly handling diverse code-mixing phenomena.

    \item \textbf{Computational Accessibility:} While fine-tuning shows substantial improvements, computational requirements and the need for substantial training data may limit accessibility for resource-constrained settings.
\end{enumerate}

\section*{Ethics Statement}

We adhere to established ethical guidelines in the creation of our benchmark dataset and in the evaluation of existing LLMs for Hinglish code-mixed text. Data curation was carried out responsibly, with careful attention to the annotator's well-being, informed consent, and workload management. We ensured that no personally identifiable information (PII) was included in the dataset, thereby maintaining user privacy and confidentiality. To mitigate potential biases, annotation protocols were designed to capture diverse linguistic phenomena and were reviewed iteratively. Our study promotes fairness and inclusivity in multilingual NLP by focusing on underrepresented code-mixed language scenarios. All datasets and models employed in this research are either publicly available or used in accordance with their respective licenses, such as Creative Commons.

\section*{Acknowledgments}
This work is supported by the Anusandhan National Research Foundation (ANRF), India, through the project titled ``Curating and Constructing Benchmarks and Development of ML Models for Low-Level NLP Tasks in Hindi-English Code-Mixing''. The authors express their gratitude to Diksha, Ronakpuri Goswami, Mahesh Kumar, Rahul Gadhvi, Yash Chopra, Mahavir Patil, Vaidahi Patel and Ashish Singh for their invaluable support with dataset annotations. We also extend our thanks to Sailesh Panda, Isha Narang and Prathamesh Shanbhag for their assistance in reveiwing the manuscript and providing feedback. Himanshu Beniwal is supported by the Prime Minister Research Fellowship (PMRF ID-1702154), India.
\bibliography{custom}

\appendix
\vspace{0pt}
\section{Appendix}
    \subsection{Dataset Sources}
    For dataset collection, we implemented an article-wise scraping process that extracted high-quality data from diverse sources.
    \textbf{News sources} included NDTV\footnote{\url{https://ndtv.in/}}, ABP News\footnote{\url{https://www.abplive.com/}}, Zee News\footnote{\url{https://zeenews.india.com/hindi}}, News18\footnote{\url{https://hindi.news18.com/}}, TV9\footnote{\url{https://www.tv9.com/}}, and Aaj Tak.\footnote{\url{https://www.aajtak.in/}} \textbf{Digital platforms} like X (formerly ``Twitter'')\footnote{\url{https://x.com/}} and YouTube\footnote{\url{https://www.youtube.com/}} provided real-time discussions. \textbf{Political channels} from INC, BJP, and AAP were included, along with \textbf{official sources} such as \textit{Mann Ki Baat}\footnote{\url{https://www.narendramodi.in/mann-ki-baat}} and \textit{Press Information Bureau (PIB)}\footnote{\url{https://pib.gov.in/}}.
    \label{sec:appendix-sources}
    
    \subsection{Examples of Noisy Text Instances in the Scrapped Code-Mixed Data}
    Table~\ref{tab:annotation-examples}  Presents examples of challenging text patterns identified during manual annotation, including incomplete variants, ambiguous scripts, cross-article concatenation, and mixed-script forms. These were carefully reviewed and, in some cases, removed as part of our annotation methodology and quality assurance process to improve dataset consistency. 
    \label{sec:appendix-examples}

\subsection{Annotation Guidelines for All Tasks}

\begin{itemize}
    \setlength{\itemsep}{0pt}
    \item Each instance was annotated independently by all annotators without influence from model predictions or other annotator's decisions.
    \item Annotators were instructed to rely on contextual understanding to disambiguate code-mixed tokens, resolve ambiguity, and accurately assign labels. 
    \item Only the content explicitly present in the sentence was to be annotated; annotators were advised to avoid adding any inferred or assumed information.  
    \item Instances containing noise (e.g., incomplete fragments, junk tokens, or malformed words) were marked and excluded during preprocessing as per filtering heuristics as per Table \ref{tab:annotation-examples}.
    \item Consistent labeling was promoted using uniform tags and task-specific instructions during annotation training.
    \item Annotators were encouraged to flag uncertain, and ambiguous samples for further review.
    \item Annotation disagreements were addressed using majority voting. In cases where no majority existed, a manual adjudication process was conducted to finalize the labels.
\end{itemize}

\textbf{Quality Control \& Training}
\begin{itemize}
\setlength{\itemsep}{0pt}
\item Annotators periodically used gold-standard examples to  ensure continued alignment throughout the annotation process.
\item Periodic sample checks provided feedback and helped uphold annotation standards.
\item An independent reviewer regularly flagged low-quality annotations for re-annotation by the original annotators.
\end{itemize}

\textbf{Conflict Resolution Strategy}
\begin{itemize}
\setlength{\itemsep}{0pt}
\item Consolidated annotation criteria: For model training and evaluation, only annotations with agreement from at least two out of three annotators were retained, ensuring reliability.
\item Iterative refinement: Disagreement patterns were analyzed to identify common sources of confusion, leading to guideline refinements and additional training for annotators.

\end{itemize}


\label{sec:guidelines}


\section{Experimental Setup}
    \label{sec:experiments} 

    \subsection*{Zero-shot LID Prompt}
    \begin{promptbox}

    You are an expert in Language Identification (LID) task for Hinglish (Hindi-English code-mixed) text. Your task is to identify and classify tokens in the given sentence.\\
    \textbf{Instructions:}\\
    - Tag each word or word group in the following text with language labels:\\
    - Use `\textcolor{orange}{hi}' for Hindi words\\
    - Use `\textcolor{blue}{en}' for English words\\
    - Use `ot' for other words\\
    - Only break tokens at spaces.\\
     Process the given sentence:\\
     \textbf{Input:} \{text\} \\
     Return the output in the following format:\\
     $[word1\; tag1,\; word2\; tag2,\; \ldots ]$

    \end{promptbox}
    
    \subsection*{One-shot LID Prompt}
    \begin{promptbox}
     You are an expert in Language Identification (LID) task for Hinglish (Hindi-English code-mixed) text. Your task is to identify and classify tokens in the given sentence.\\
     \\
     \textbf{Instructions:} \\
     - Tag each word or word group in the following text with language labels:\\
     - Use `\textcolor{orange}{hi}' for Hindi words (e.g., \textcolor{orange}{Mujhe}, \indicWords{agsentence}, \textcolor{orange}{karna}, \indicWords{aksentence}, \textcolor{orange}{hai}, \indicWords{alsentence}, \textcolor{orange}{shala})\\
     - Use `\textcolor{blue}{en}' for English words (e.g., \textcolor{blue}{Culture}, \indicWords{amsentence}, \textcolor{blue}{Lifestyle}, \textcolor{blue}{for}, \indicWords{ansentence}, \textcolor{blue}{Alliance}, \indicWords{aosentence}, \textcolor{blue}{of}, \textcolor{blue}{initiative}, \indicWords{apsentence})\\
     - Use `ot' for other words (e.g., \#Bollywood, \#BJP, @PMOIndia, . - : @ = \& * +)

      - Be precise and consistent with tags classification. \\
      - Do not add any other extra suggestions. \\
      - Only break tokens at spaces. \\
      - Format: Return space-separated word-tag pairs. 

\textbf{Example Input:} \indicWords{Wsentence}\indicWords{Xsentence}\indicWords{Ysentence} \textcolor{blue}{$21^{st}$ Commonwealth Games} \indicWords{Zsentence} \textcolor{blue}{India} \indicWords{asentence} \indicWords{bsentence} \textcolor{blue}{first Gold medal} \indicWords{csentence}\indicWords{dsentence} |

\textbf{Output}: 
\indicWords{Wsentence} \textcolor{orange}{hi} \indicWords{Xsentence} \textcolor{orange}{hi} \indicWords{Ysentence} \textcolor{orange}{hi} {$21^{st}$} ot \textcolor{blue}{Commonwealth en Games en} \indicWords{Zsentence} \textcolor{orange}{hi} \textcolor{blue}{India} \textcolor{blue}{en} \indicWords{asentence} \textcolor{orange}{hi} \indicWords{bsentence} \textcolor{orange}{hi} \textcolor{blue}{first en Gold en medal en} \indicWords{csentence} \textcolor{orange}{hi} \indicWords{dsentence} \textcolor{orange}{hi} | ot 

\textbf{Input:} \{text\}
    \end{promptbox}
    \subsection*{Zero-shot MLI Prompt}
   
    \begin{promptbox}
    You are a helpful AI Assistant and your task is to identify and determine the \textbf{matrix language (dominant grammatical structure language)} of the given Hinglish (Hindi-English) code-mixed sentence.\\
    \\
    \textbf{Instructions:} \\
    The matrix language is the main language that governs the grammatical structure of the sentence. It may borrow words from another language, but the syntax and morphology will mostly follow the matrix language.\\ 
    Process the given sentence:\\
    \textbf{Input:} \{text\} \\
    Return the output in the following format: only the matrix language name. 
    \end{promptbox}

    \subsection*{One-shot MLI Prompt}
    \begin{promptbox}
    Your task is to identify and determine the matrix language (dominant grammatical structure language) of the given Hinglish (Hindi-English) code-mixed sentence.\\
    \\
    The matrix language is the main language that governs the grammatical structure of the sentence. It may borrow words from another language, but the syntax and morphology will mostly follow the matrix language.\\
    \\
    \textbf{Instructions:}\\
    1. If the sentence is primarily structured in Hindi, respond with: `\textcolor{orange}{hi}'\\
    2. If the sentence is primarily structured in English, respond with: `\textcolor{blue}{en}'\\
    3. Respond with a single word only: `\textcolor{orange}{hi}' or `\textcolor{blue}{en}'. Do not add any other extra suggestions.\\
    \\
    \textbf{Example Input:} \textcolor{blue}{India's automation and design expert pool is vast}, \indicWords{Qsentence}\textcolor{blue}{Global} \indicWords{Rsentence} |

    \textbf{Output:} \textcolor{blue}{en} \\

    Process the given sentence:\\
    \textbf{Input:} \{text\}  
    \end{promptbox}
    \subsection*{Zero-shot NER Prompt}
    \begin{promptbox}
    You are a helpful AI Assistant and your task is to identify the named entities in the following Hinglish code-mixed sentence.\\
    \textbf{Instructions:}\\
    - Tag each word with one of these entity types:\\
      PERSON, ORGANISATION, LOCATION, DATE, TIME, GPE, HASHTAG, EMOJI, MENTION, X - for all other words. \\
      \\
      Process the given sentence:\\
    \textbf{Input:} \{text\}
    \\
    Return the output in the following format:\\
    $[$ $\{$ `word1': `entity'$\}$, $\{$ `word2': `entity' $\}$, $\{$ `word3': `entity' $\}$, $\ldots$ $]$

    \end{promptbox}
    
    \textbf{One-shot NER Prompt}
    \begin{promptbox}
     You are a helpful AI Assistant and your task is to identify the named entities in the following Hinglish code-mixed sentence.\\
     \\
     \textbf{Instructions:} \\
     - Tag each word with one of these entity types:\\
     PERSON - for names of people\\
     ORGANISATION - for organization names\\
     LOCATION - for location names\\
     DATE - for dates\\
     TIME - for time expressions\\
     GPE - for geo-political entities\\
     HASHTAG - for words starting with \#\\
     EMOJI - for emojis\\
     MENTION - for words starting with @\\
    X - for words that don't fall into above categories.\\
    - Only break tokens at spaces.\\
    -  Do not add any extra explanations or text before or after the list. 


\textbf{Example:} \indicWords{msentence} \indicWords{afsentence}\textcolor{blue}{Madame Tussauds} \indicWords{acsentence} \textcolor{blue}{Deepika Padukone} \indicWords{adsentence}\indicWords{agsentence} \indicWords{aesentence}|

\textbf{Output}: [ \indicWords{msentence} GPE, \indicWords{afsentence} X, \textcolor{blue}{Madame} LOCATION, \textcolor{blue}{Tussauds} LOCATION, \indicWords{acsentence} X, \textcolor{blue}{Deepika} PERSON, \textcolor{blue}{Padukone} PERSON, \indicWords{afsentence} X, \indicWords{ausentence} X, \indicWords{avsentence} X, \indicWords{awsentence} X, \indicWords{agsentence} DATE, \indicWords{azsentence}X, \indicWords{axsentence}X, \indicWords{aysentence}X, | X ]

\textbf{Input:} \{text\}
\end{promptbox}
    
    \subsection*{Zero-shot POS Tagging Prompt}
    \begin{promptbox}
    Your task is to assign Part-of-Speech (POS) tags to each word or word group in the given code-mixed sentence.\\
    \\
    \textbf{Instructions:}\\
    - Tag each word with the appropriate grammatical category from the provided tagset.\\
    - Use available POS tags: VERB, NOUN, PRON, ADJ, ADV, ADP, PROPN, CONJ, DET, NUM, PART, PRON\_WH, PART\_NEG, X.\\
    - Only break tokens at spaces.\\ 
    \\
    Process the given sentence:\\
    \textbf{Input:} \{text\}\\
    Return the output in the following format: [[\{`word1': `POS\_TAG1', `word2': `POS\_TAG2'\}, \{'word3': `POS\_TAG3', `word4': `POS\_TAG4'\}, ...]]\\
    \end{promptbox}
    
    
    \subsection*{One-shot POS Tagging Prompt}
    \begin{promptbox}
    You are a linguistics expert specializing in Part-of-Speech (POS) tagging, particularly for code-mixed Hindi-English (Hinglish) text.\\
    Given a Hinglish sentence, provide a token-wise POS tag for each word in JSON format. Ensure accurate tagging for both Hindi and English words, considering the context and mixed grammar structures. \\
 
    \textbf{Instructions:} \\
    1. Analyze each word in the sentence and identify the correct POS tag. \\
    2. Be precise and consistent with POS classification. \\
    3. Consider the grammatical context of code-mixed structures. \\    4. Do not add any other extra suggestions. \\
    5. Use only the following tagset: \\
    - NOUN: Common nouns \\
    - PROPN: Proper nouns  \\
    - VERB: Verbs in all forms \\
    - ADJ: Adjectives \\
     \end{promptbox}
     \begin{promptbox}
    - ADV: Adverbs \\
    - ADP: Adpositions (pre/postpositions) \\
    - PRON: Pronouns \\
    - DET: Determiners \\
    - CONJ: Conjunctions \\
    - PART: Particles \\
    - PRON\_WH: Question words \\
    - PART\_NEG: Negation words \\
    - NUM: Numbers \\
    - X: Other (punctuation, foreign words) \\

   \textbf{Example Input:} \indicWords{Wsentence}\indicWords{Xsentence}\indicWords{Ysentence} \textcolor{blue}{$21^{st}$ Commonwealth Games} \indicWords{Zsentence} \textcolor{blue}{India} \indicWords{asentence} \indicWords{bsentence} \textcolor{blue}{first Gold medal} \indicWords{csentence}\indicWords{dsentence} |

    \textbf{Output:} [`\indicWords{Wsentence}': `PROPN', `\indicWords{Xsentence}': `PROPN', `\indicWords{Ysentence}': `PART', `$21^{st}$': `NUM', `\textcolor{blue}{Commonwealth}': `PROPN', `\textcolor{blue}{Games}': `PROPN', `\indicWords{Zsentence}': `ADP', `\textcolor{blue}{India}': `PROPN', `\indicWords{asentence}': `ADP', `\indicWords{bsentence}': `ADP', `first': `ADJ', `\textcolor{blue}{Gold}': `NOUN', `\textcolor{blue}{medal}': `NOUN', `\indicWords{csentence}': `VERB', `\indicWords{dsentence}': `VERB', `|': `X']

     \textbf{Input:} \{text\}
    \end{promptbox}

\label{sec:appendix-prompts}

\subsection*{Zero-shot MT Prompt}
\begin{promptbox}
You are a helpful AI Assistant specializing in machine translation for code-mixed Hindi-English (Hinglish) text. Your task is to translate Hinglish sentences into three different formats while maintaining meaning and natural flow. \\

Given a Hinglish code-mixed sentence, provide translations in the following three formats:

1. \textbf{Standard English}: Complete fluent and grammatically correct English translation\\
2. \textbf{Romanized Hindi}: Complete translation in Hindi using Roman/Latin script\\  
3. \textbf{Devanagari Hindi}: Complete translation in fluent Hindi using Devanagari script\\

Process the given sentence:\\
\textbf{Input:} \{text\}\\
Return the output in the following format:\\
English: [English translation]\\
Romanized Hindi: [Hindi in Roman script]\\
Devanagari Hindi: [Hindi in Devanagari script]
\end{promptbox}

\subsection*{One-shot MT Prompt}
\begin{promptbox}
You are a helpful AI Assistant specializing in machine translation for code-mixed Hindi-English (Hinglish) text. Your task is to translate Hinglish sentences into three different formats while maintaining meaning and natural flow.

Given a Hinglish code-mixed sentence, provide translations in the following three formats:\\
1. \textbf{Standard English}: Complete fluent and grammatically correct English translation\\
2. \textbf{Romanized Hindi}: Complete translation in Hindi using Roman/Latin script\\  
3. \textbf{Devanagari Hindi}: Complete translation in fluent Hindi using Devanagari script

\textbf{Instructions:}\\
1. Ensure all translations convey the same meaning as the original Hinglish text.\\
2. Maintain natural flow and grammatical correctness in each target format.\\
3. Consider cultural context and idiomatic expressions appropriately.\\
4. Do not add any other extra suggestions or explanations.\\

\textbf{Example Input:} \indicWords{qsentence}{\color{blue}INDIAN NAVY}\indicWords{rsentence}\indicWords{ssentence}\indicWords{tsentence}\indicWords{usentence}\indicWords{vsentence}\indicWords{wsentence} \indicWords{xsentence}\indicWords{ysentence}\indicWords{zsentence}|

\textbf{Output:}\\
English Translation: {\color{blue}The Indian Navy headquarters located in Delhi and the Western Naval Command have jointly orchestrated the formation of the operation.}\\
Romanized Hindi Translation: {\color{orange}Dilli sthit }{\color{blue}Indian Navy headquarters} {\color{orange}aur} {\color{blue}Western Naval Command} {\color{orange}ne milkar} {\color{blue}operation} {\color{orange}ke vyuh ki rachna ki hai} .\\
Devanagari Hindi Translation: \indicWords{qsentence}
\indicWords{aasentence}\indicWords{atsentence}\indicWords{ssentence}\indicWords{tsentence}\indicWords{usentence}\indicWords{vsentence} \indicWords{wsentence}\indicWords{xsentence}\indicWords{ysentence}\indicWords{zsentence}|

Process the given sentence:\\
\textbf{Input:} \{text\}
\end{promptbox}

\subsection{Fine-Tuning Hyperparameters}
\label{sec:hyperparameters}

The optimization process focused on fine-tuning four key hyperparameters, with the goal of balancing the refinement of essential parameters while minimizing unnecessary adjustments to those already well-suited for the task. Batch size, number of epochs, weight decay, and learning rate were selected due to their direct and substantial impact on model performance, stability, and generalization.

\noindent \textbf{Core Training Parameters:} \\
\textbf{Epochs:} 3 \\
 \textbf{Batch size:} 4 per device with gradient accumulation steps of 8 (effective batch size: 32) \\
\textbf{Learning rate:} 2e-4 with cosine scheduler and warmup ratio of 0.1 \\
\textbf{Weight decay:} 0.01 \\
\textbf{LoRA Configuration:} LoRA with rank 32, alpha 64, dropout 0.1.

\noindent \textbf{Instruction Format:} All tasks used task-specific instruction templates with examples, following the format: \\
\texttt{Instruction: [task description] \\ 
Sentence: [input] \\
Output: [expected output]}
\section{Computation Requirement and Budget}

The experiments were conducted using API-based access to state-of-the-art Large Language Models (LLMs), including \texttt{gpt-4o}, Command R+ (\texttt{command-a-03-2025}) by Cohere, and  \texttt{claude-3.5-sonnet}. The estimated monthly costs for API usage were approximately \$200 for \texttt{claude-3.5-sonnet}, \$150 for Cohere, and \$50 for \texttt{gpt-4o}, resulting in a total estimated cost of \textbf{\$400} per month. For computational infrastructure, experiments were carried out on four NVIDIA Tesla V100 32 GB GPUs, with an estimated cost of \$7,192.00 per month based on Google Cloud Platform (GCP) \footnote{\url{https://cloud.google.com/products/calculator}} Calculator pricing.
\label{sec:compute}

\begin{table*}[!htbp]
\centering
\small
\begin{tabular}{p{0.30\linewidth}p{0.6\linewidth}}
\toprule
\textbf{Category} & \textbf{Example Text} \\
\midrule
Incomplete variant & ), {\color{blue}floppy disk, hard disk drive, magnetic stripe card, relational database, SQL} \indicWords{csentence}  ({\color{blue}DRAM}) ({\color{blue}Dynamic Random-Access Memory}) \indicWords{dsentence} \\
\midrule
Ambiguous script & {{Menu}}<br/>\indicWords{esentence}.jpg|thumb]] ===++ Image \indicWords{fsentence} / [[:en:{\color{blue}Giridhar Lal Aggarwal Freedom Fighter} | {\color{blue}Giridhar Lal Aggarwal}]] == |\\
\midrule
Cross-article concatenation  & \indicWords{gsentence}[.....] 08/10/2020 {\color{blue}{Satyam KushwahLeave a Comment on}} \indicWords{hsentence}| \\
\midrule
Mixed-script variant & @Strawberigloz he barobar naahi aahe, aaplich manasa aaplyala paathi sodtat. Aaplya itithasacha garva asla pahije. \\
\bottomrule
\end{tabular}
\caption{Examples of noisy text instances in the dataset containing mixed content and transitions. \textbf{\textit{Takeaway}}: These noisy text instances in the dataset reflect challenges in code-mixed annotation, require careful preprocessing.}
\label{tab:annotation-examples}
\end{table*}

\vspace{-1em}

\begin{table*}[!htbp]
\centering
\small
\setlength{\tabcolsep}{0.35em}
\renewcommand{\arraystretch}{1.2}
\setlength{\tabcolsep}{0.35em}
\renewcommand{\arraystretch}{1.2} 
\begin{tabular}{>{\raggedright\arraybackslash}p{1.0cm} 
                >{\centering\arraybackslash}p{6.2cm} 
                >{\centering\arraybackslash}p{1.9cm} 
                >{\centering\arraybackslash}p{1.3cm} 
                >{\centering\arraybackslash}p{1.2cm} 
                >{\centering\arraybackslash}p{2.7cm}}
\toprule
\textbf{Task} & \textbf{Data Source (Hi-En)} & \textbf{Dataset Size } & \textbf{Script} & \textbf{QA} & \textbf{Annotators/Models} \\
\midrule
\multirow{7}{*}{\begin{sideways}LID\end{sideways}}
& Facebook \citep{bali-etal-2014-borrowing} & 1,062 & R \& D & Yes & 3 \\
& Twitter \citep{singh-etal-2018-language} & 2,079 & R & Yes & 3 \\
& Twitter \citep{swami2018corpus} & 5,250 & R & Yes & Not mentioned \\
& Twitter \citep{mave-etal-2018-language} & 5,567 & R & Yes & 3 \\
& Facebook, Twitter, WhatsApp \citep{veena2018character} & 3,071 & R & No & Embedding Model \\
& Twitter \citep{joshi2022evaluating} & 18,461 & R & No & Not mentioned \\
& \textbf{Twitter, YouTube, Press Releases, News (Ours)} & \textbf{25,773} & \textbf{R \& D} & \textbf{Yes} & \textbf{3} \\
\midrule
\multirow{4}{*}{\begin{sideways}MLI\end{sideways}}
& Twitter, Facebook \citep{sequiera-etal-2015-pos} & 628 & R \& D & No & 1 \\
& Facebook \citep{bali-etal-2014-borrowing} & 1,062 & R \& D & Yes & 3 \\
& Social Media \citep{dhar-etal-2018-enabling} & 6,096 & R & Yes & 4 \\
& \textbf{Twitter, YouTube, Press Releases, News (Ours)} & \textbf{25,773} & \textbf{R \& D} & \textbf{Yes} & \textbf{3} \\
\midrule
\multirow{6}{*}{\begin{sideways}NER\end{sideways}}
& Facebook \citep{bali-etal-2014-borrowing} & 1,062 & R \& D & Yes & 3 \\
& Twitter \citep{singh-etal-2018-language} & 2,079 & R & Yes & 3 \\
& Twitter \citep{bhargava2016named} & 2,700 & R & No & Supervised algorithm \\
& Twitter \citep{singh-etal-2018-named} & 3,638 & R & Yes & 2 \\
& Tourism, News \citep{murthy2022hiner} & 108,608 & R \& D & No & 1 \\
& \textbf{Twitter, YouTube, Press Releases, News (Ours)} & \textbf{24,913} & \textbf{R \& D} & \textbf{Yes} & \textbf{3} \\
\midrule
\multirow{7}{*}{\begin{sideways}POS \end{sideways}}
& Twitter, Facebook \citep{sequiera-etal-2015-pos} & 628 & R \& D & No & 1 \\
& Facebook \citep{bali-etal-2014-borrowing} & 1,062 & R \& D & Yes & 3 \\
& Twitter, Facebook \citep{jamatia2015part} & 1,106 & R & No & 2 \\
& Twitter \citep{singh-etal-2018-twitter} & 1,190 & R & Yes & 3 \\
& Synthetically generated \citep{chatterjee2022pacman} & 51,118 & R \& D & No & 0 \\
& Existing Benchmarks \citep{kodali-etal-2022-symcom} & 55,474 & R & No & Trained POS tagger \\
& \textbf{Twitter, YouTube, Press Releases, News (Ours)} & \textbf{24,598} & \textbf{R \& D} & \textbf{Yes} & \textbf{3} \\
\midrule
\multirow{6}{*}{\begin{sideways}MT\end{sideways}}
& TED Talks, News, Wikipedia~\cite{kartik2024synthetic} & 2,787 & R \& D & Yes & 2 \\
& Twitter, Facebook~\cite{srivastava-singh-2021-hinge} & 3,952 & R \& D & Yes & 5 \\
& Social Media~\cite{dhar-etal-2018-enabling} & 6,096 & R & Yes & 4 \\
& Twitter, Facebook~\cite{srivastava-singh-2020-phinc} & 13,738 & R & Yes & 54 (400 instances) \\
& Existing Benchmarks~\cite{kunchukuttan2017iit} & 14,95,854 & R \& D & No & PBSMT, NMT \\
& \textbf{Twitter, YouTube, Press Releases, News (Ours)} & \textbf{24,558} & \textbf{R \& D} & \textbf{Yes} & \textbf{2} \\
\bottomrule
\end{tabular}
\caption{Comprehensive Comparison of Existing Datasets for Hinglish Code-Mixing NLP Tasks, including the proposed dataset. NLP tasks covered in the dataset include Language Identification (LID), Part-of-speech (POS) tagging, Named Entity Recognition (NER), Matrix Language Identification (MLI) and Machine Translation (MT). (R) and (D) denote Roman and Devanagari scripts, respectively, while QA represents annotations by Qualified Annotators. 
}
\label{tab:dataset_comparison}
\end{table*}

\begin{table*}[!htbp]
\centering
\begin{tabular}{p{0.30\linewidth}p{0.6\linewidth}}
\toprule
\textbf{Response Flaw Type} & \textbf{Example Behavior or Observation} \\
\midrule
Script and entity Misidentification 
& Words such as `\indicWords{msentence}', which are borrowed English terms written in Devanagari, are frequently misclassified as Hindi by most models. Additionally, models like \texttt{gpt-4o} demonstrate entity misclassification issues, such as tagging `\textcolor{blue}{Union Home Minister}' as an \textsc{ORGANISATION} and `\textcolor{blue}{Holi}' as a \textsc{DATE}. \\
\midrule
Sentence Truncation & Long-form code-mixed inputs lead to abrupt endings or incomplete generations (e.g., output stops mid-sentence despite ample context). \\
& In `\textcolor{blue}{Yes, we belong to this place} – \indicWords{ahsentence} \indicWords{aisentence}', only `\textcolor{blue}{Yes, we belong to this place}' is translated to (\indicWords{ajsentence}/ \textcolor{orange}{haan, hum is jagah se sambandhit hain}), while the rest is ignored despite clear context.\\
\midrule
Repetitive Generation & Models like \texttt{gemini-1.5-flash} and \texttt{mistral-instruct} frequently exhibit repetitive generation patterns. For instance, they may produce outputs such as: `\textcolor{blue}{The second tagging is more accurate as it identifies `this' as a determiner and `last' as a quantity}'. repeating similar explanations or sentence fragments within the same response.
 \\
\midrule
Subjective Additions & Instead of remaining factual, models add speculative commentary (e.g., `\textcolor{blue}{en: The given text is in English. The hashtag} `\#MadeByGoogle' \textcolor{blue}{is also in English.  `E' (English).}'). 
\\
\midrule
Prompt Mimicry & \texttt{gpt-4o} and \texttt{command-r-plus} mirror example formats from the prompt, failing to adapt to new inputs and instead mimicking example structure. `\textcolor{blue}{Based on the given text, it is written in the Hindi language. Therefore, the matrix language label for this sentence is h}'.\\
\midrule
High-variance Failure & Inputs with abrupt transitions, broken grammar, or inconsistent scripts result in empty, irrelevant, or default responses. \\
& \textbf{Example 1:} lakhanOo: \textcolor{blue}{dr. apj abdul kalam} bhArat kA 11veN rAShTrapati thE karoDhON bhAratiyON ke lIyE prEraNAdhA kA strOt thE, dR. apj abdul kalam \\
& \textbf{Example 2:} This text does not contain any GPE, DATE, TIME, HASHTAG, EMOJI, or MENTION entities. \\
\midrule
Hallucination &  Models like \texttt{mistral-instruct} fabricate non-existent locations or attributes (e.g., inventing MATCH, VERSION, COUNTRY, PRODUCT, QUANTITY, or BUILDING categories not present in the input). \\
& \textbf{Example 1:} Note: The context is assumed to be empty in this example. If context information is available, it should be provided to improve the accuracy of entity tagging. \\
& \textbf{Example 2:} ḍipṭī sīṭī mānejar lēmūēla rāṇḍolpha nē kahā, 'splāś pāḍ kā rōjānā sāf-sāfāī kamī huī hai \\
\bottomrule
\end{tabular}
\caption{Observed limitations across LLMs while processing noisy, code-mixed text. \textbf{\textit{Takeaway}}: Failures are diverse - ranging from linguistic issues to structural hallucinations and prompt sensitivity - highlighting the need for integrated data-centric training strategies that can effectively handle linguistic and structural complexities.}
\label{tab:llm-failure-cases}
\end{table*}


\begin{table*}[!htbp]
\centering
\begin{tabular}{p{0.30\linewidth}p{0.6\linewidth}}
\toprule
\textbf{Response Flaw Type} & \textbf{Example Behavior or Observation} \\
\midrule
Output Blank or Missing 
& The \texttt{Qwen2.5-7B-Instruct} and \texttt{LLaMA-3.1-8B-Instruct} models frequently returned blank or no output for complex sentences across all tasks, particularly in zero-shot settings when encountering noisy or script-mixed inputs. \\
\midrule
Task Instruction Violation & Models like \texttt{aya-expanse-8b} and \texttt{LLaMA-3.1-8B-Instruct} generated Python code snippets in some task outputs: 
\newline
`\# Split the input text into words based on spaces
\newline
\texttt{words = text.split()} 
\# Initialize an empty list for words
\texttt{tagged\_words = []} \# Iterate over each word in input text given  \texttt{for word in words:} \# Check if the word is in the Hindi' for LID tasks, failing to follow prompt formatting requirements and giving incorrect output. \\
\midrule
Subjective Commentary Injection & All Models added unnecessary disclaimers like `\textcolor{blue}{Please note that the output is a best-effort attempt and might not be 100\% accurate due to the complexity of Hinglish language}' and `\textcolor{blue}{Please let me know if you need any further assistance}!' instead of providing direct outputs. \\
\midrule
Language Script Confusion & \texttt{aya-expanse-8b} exhibited severe input-output disconnection and incorrect script identification in MT and MLI tasks. For MT, an input about \textcolor{blue}{Bangladesh football team} yielded an unrelated output discussing Republic TV and Arnab Goswami in ``Urdu'' script. For MLI, the input `\indicWords{aqsentence}' was misclassified as ``Bengali (Bangla)'' instead of Hindi, indicating a failure to correctly identify the matrix language.\\
\midrule
Entity Type Misclassification & All Models frequently misclassified entities in code-mixed contexts, such as tagging `\textcolor{blue}{Uttar Pradesh}' as separate mismatched entities (`\textcolor{blue}{Uttar}':ORGANISATION, `\textcolor{blue}{Pradesh}':PLACE) instead of the correct unified LOCATION labels (`\textcolor{blue}{Uttar}':LOCATION, `\textcolor{blue}{Pradesh}':LOCATION), demonstrating poor understanding of entity boundaries. \\
\midrule
Hallucination & \texttt{aya-expanse-8b} in one-shot POS tagging created its own input sentence
(`\indicWords{arsentence}') 
different from the actual input and monolingual, then tagged the fabricated sentence. \\
\midrule
Inconsistent Output Formats & \texttt{LLaMA-3.1-8B-Instruct} and \texttt{Qwen2.5-7B-Instruct} provided inconsistent MLI labels like `Mixed', `Code-mixed language: Hindi-English' instead of `\textcolor{orange}{hi}' or `\textcolor{blue}{en}', for some inputs, showing format instability across evaluation instances. \\
\midrule
Multi-language Script Errors & For zero-shot MT, \texttt{Qwen2.5-7B-Instruct} generated Romanized Hindi in Arabic-Urdu, Bangla-English script instead of Latin script, completely misunderstanding and hallucinating in some input instances. \\
\bottomrule
\end{tabular}
\caption{Observed limitations across 7--8B parameter LLMs during zero-and one-shot evaluation. \textbf{\textit{Takeaway}}: While smaller models exhibit severe failure patterns in zero-shot and one-shot settings—fine-tuning on code-mixed data transforms them into highly capable systems that often match or exceed larger proprietary models, demonstrating the importance of task-specific training for deploying compact models in multilingual scenarios.}
\label{tab:8b-model-failures}
\end{table*}

\FloatBarrier

\end{document}